# Biometric Matching and Fusion System for Fingerprints from Non-Distal Phalanges

Mehmet Kayaoglu[1], Berkay Topcu[2], Umut Uludag[2]

[1]Bahcesehir University, Turkey
[2]TUBITAK – BILGEM, Informatics and Information Security Research Center, Turkey
mehmet.kayaoglu@stu.bahcesehir.edu.tr, {berkay.topcu, umut.uludag}@tubitak.gov.tr
www.ekds.gov.tr/bio

***Abstract*** *Market research indicates that fingerprints are still the most popular biometric modality for personal authentication. Even with the onset of new modalities (e.g. vein matching), many applications within different domains (e-ID, banking, border control…) and geographies rely on fingerprints obtained from the distal phalanges (a.k.a. sections, digits) of the human hand structure. Motivated by the problem of poor quality distal fingerprint images affecting a non-trivial portion of the population (which decreases associated authentication accuracy), we designed and tested a multifinger, multiphalanx fusion scheme, that combines minutiae matching scores originating from non-distal (ie. middle and proximal) phalanges based on (i) simple sum fusion, (ii) NFIQ image-quality-based fusion, and (iii) phalanx-type-based fusion. Utilizing a medium-size (50 individuals, 400 unique fingers, 1600 distinct images) database collected in our laboratory with a commercial optical fingerprint sensor, and a commercial minutiae extractor & matcher (without any modification), allowed us to simulate a real-world fingerprint authentication setting. Detailed analyses including ROC curves with statistical confidence intervals show that the proposed system can be a viable alternative for cases where (i) distal phalanx images are not usable (e.g. due to missing digits, or low quality finger surface due to manual labor), and (ii) switching to a new biometric modality (e.g. iris) is not possible due to economical or infrastructure limits. Further, we show that when distal phalanx images are in fact usable, combining them with images from other phalanges increases accuracy as well.*

## 1. Introduction

Biometrics-based personal authentication systems are increasingly replacing or augmenting token (e.g. ID card) and/or knowledge (e.g. username & password) based systems. These systems are utilizing physiological and/or behavioral characteristics such as fingerprint, face, iris, voice, hand geometry for personal authentication. As they provide several important advantages over traditional authentication mechanisms, such as user convenience and increased security, the number of biometric installations and the affected public sizes are increasing rapidly, with applications ranging from national e-ID programs to bank ATM access, from critical infrastructure security to mobile authentication. With projected further increases in (i) world population, and percentage of people living in close proximity in cities & associated security concerns, (ii) mobile/internet-based access for many services (e.g., education, standardized exams, payments, commerce, voting), (iii) businesses shifting their practices into online realm (e.g., proliferation of home-office, e-office options), and (iv) cloud-based IT services which require highly secure & accurate authentication sub-systems for protecting critical data, biometrics will continue to be an important research / application / evaluation domain for academia, industry, and governments around the world.

Among many biometric modalities, fingerprints are the most popular choice for many reasons listed below: (i) fingerprint sensors are –generally– cheaper & smaller & they need less power to operate than other biometric sensors (e.g., hand geometry), hence, they are especially suitable for mobile environments with limited space / energy resources and applications with budgetary restrictions, (ii) as it has been commercialized earlier, fingerprint technology enjoys the presence of standards for features used for matching (e.g., ISO 19794-2 [1]), which increase sensor / algorithm / system interoperability and allow widespread deployments, across different applications and geographies, (iii) the authentication accuracy characterized by Genuine Accept Rates and False Reject Rates (abbreviated as GAR and FRR, respectively) for fingerprints is among the "best performing" set (along with iris matching [2]), (iv) there are many standardized tests (e.g., FVC series [3-5], MINEX [6], FpVTE [7]), that help decision makers in government / industry / academia, in objectively & quickly judging the performance tradeoffs for different sensors / algorithms / systems, (v) there are many public fingerprint databases (e.g., [3-5]) available to researchers that help in obtaining fair comparisons of algorithm performances, (vi) existing – mainly– fingerprint-based AFIS/ABIS infrastructure of law-enforcement agencies (e.g., FBI, police, homeland & border security) makes fingerprint-based authentication a logical & cost-effective choice.

As the single biometric source in associated *unimodal* systems, or as an important module in *multimodal* systems that combine many biometric sources (e.g., face, iris and fingerprint [8]) to improve accuracy & user friendliness, fingerprint-based authentication works typically with images (or, features extracted) from the distal *phalanges* (a.k.a. sections, digits) of the human finger structure. Majority of these systems utilize *minutiae*-based features (ie. characteristics for the discontinuities in the regular ridge flows on the associated surfaces), and to a lesser extent, *texture* features (such as Gabor filter responses) for comparing two fingerprint images and arriving at matching scores [9].

Unfortunately, for a non-trivial portion of subject databases and across different applications, it is found that, (i) some (distal) fingerprints are, naturally, of very low quality to be useful in biometric matching, regardless of the sensor technology utilized (e.g., capacitive, thermal, optical), (ii) environmental factors such as temperature, humidity, wear due to manual labor etc. affect the image (and also, associated template) quality dynamically & sometimes drastically.

In this paper, firstly, we report fingerprint minutiae matching results for non-distal (ie. middle and proximal) phalanges of fingers from a mid-size database –composed of 50 individuals, 400 unique fingers, 1600 distinct images– we collected in our laboratory. Originating from our earlier work [10, 11], where we introduced the individual matching accuracy comparisons and single-finger fusion results for different phalanges, in this paper we then extend our analyses to *multifinger & multiphalanx* fusion (based on three fusion techniques: simple sum, NFIQ image-quality- based, and phalanx-type-based, where we found that the last one provides the highest accuracy), and provide detailed statistics regarding (i) measured image quality (utilizing popular NFIQ metric [12]), (ii) minutiae count, and (iii) authentication accuracies with statistical confidence intervals.

As our results indicate, multifinger (e.g. two fingers) fusion of sequentially obtained non-distal phalanges lead to authentication accuracies exceeding those associated with traditionally used distal phalanges. Hence, we show that non-distal fingerprint phalanges may provide an accurate verification source, in cases where traditionally utilized distal phalanges cannot be used (e.g. due to low quality finger surfaces or missing digits). Note that, we are certainly not implying that non-distal phalanges should replace the -traditional- distal ones as the biometric modality to use in a generic setting. Rather, we are just showing an alternative route, supported with quantitative experimental results, which can be logically utilized if (i) the traditional distal phalanges are not usable, and (ii) switching to a completely new biometric modality (e.g. iris, finger vein, palm print) is not possible due to cost and other application-related (e.g. new sensor deployment, manufacturing design …) limits.

Since we have utilized a commercial & low-cost optical fingerprint sensor and a commercial minutiae feature extractor / matcher in our studies, without any modification at all, we basically simulate an environment where we can measure the performance of an existing, traditional (distal phalanx-based) fingerprint matching system against the performance of the proposed fusion system (without changing the sensor & feature extractor / matcher). It is quite logical to state that, if a feature extractor / matcher tuned specifically to non-distal phalanx characteristics were used, the associated authentication performances would be even higher than what we report in this manuscript. But, we wanted to compare non-distal and distal phalanges, *without* changing the associated algorithmic infrastructure at all. We believe that this choice increases the applicability of the proposed system in real-world commercial / governmental application scenarios.

Note that, our fusion scheme can be classified as a *multibiometric* [13] system, where, we combine features from a single modality, but using multiple signal acquisitions (of different portions of the finger structure) via a single sensor. Compared to a *multimodal* system (e.g., combining iris and fingerprint), where one has to operate multiple sensors & associated extractor / matcher pairs, our system can be more economical and easier to use. But, we should also note that, a multimodal system, where intrinsically independent biometric sources (e.g., iris and fingerprint) are fused, can benefit from this independence: when one source is not (relatively) useful for matching, the other may be. Hence, compared to fusing relatively dependent sources (e.g., fusing matching scores from index finger & middle finger, or fusing thumb print with palmprint, etc.), multimodal fusion can enjoy higher accuracies, at the cost of increased sensor / extractor / matcher system costs.

The proposed fusion system basically utilizes the existing fingerprint sensor / algorithm infrastructure, and adaptively fuses the individual matching scores originating from different phalanges based on (i) simple-sum, (ii) image quality, and (iii) phalanx-type. To the best of our knowledge, our analysis is the first one dealing with fingerprint phalanges' authentication data fusion, for commercial (unmodified) fingerprint sensor / feature extractor / matcher paradigm.

Previously, Ribaric and Pavesic [14] utilized an image scanner (operating at 600 DPI) to capture finger and digit prints. Using a texture-based feature extractor and fusion of two digits, they arrived at approximately 98% correct identification rate. Results of similar studies utilizing image scanners can be found in [15,16]. A study by Zhang *et al*. [17] deals with another region of the finger structure other than the distal phalanx, namely, the knuckle on the outer surface of the joint between the phalanges. The authors utilized Gabor texture features on images acquired with a custom-built sensor. Another texture-based knuckle print recognition algorithm is proposed in [18].

In [19], authors use a webcam to image the palm of the hand, and use ridgelet transform for processing images of phalanges. Rowe *et al*. [20] utilized a large, prototype whole-hand multispectral imager, to capture palmprint and all finger phalanges' images in a single impression. Using score-level fusion with these sources, they were able to arrive at near-perfect genuine / imposter score distribution separations. Uhl and Wild [21] analyzed several hand-based biometric features (e.g., minutiae, eigenfingers from digits, palmprint, hand shape) for assessing the authentication accuracy differences between children and adult users. They have utilized an image scanner (operating at 500 DPI) to collect their experimental data, and minutiae matches were confined to distal phalanges. In [22], second minor finger knuckle features are used for image-based matching.

A recent demonstration by Yahoo Labs researchers [23] utilized smart phone touch screens for capturing images of the palm, ear, and finger phalanges for personal authentication.

As a more general note, for an introduction to statistical classifier combination & theoretical evaluation of different fusion techniques, the reader is referred to the comprehensive work in [24]. In [25], authors summarize a general framework for quality-based biometric fusion, not specifically pertaining to finger biometrics, but applicable to any combination of biometric signals. In [26,27], not specifically considering biometrics, but dealing with generic classifier combination, the authors evaluate different criteria (ie. independence and diversity measures) in terms of resulting classifier performance. Kittler et al. [28], experimentally shows (and provides a theoretical explanation for) the superior performance of sum-rule, as compared to max, min, median and majority vote fusion rules: the authors state that the increased resilience of sum rule against estimation errors is a possible factor contributing to its success.

Note that, in this manuscript we also evaluated the sum rule for fusing scores from different fingerprint phalanges, and found that simple-sum fusion achieves good performance, reaching the levels attained by image-quality-based fusion. However, the third fusion rule that we utilized (phalanx-type-based fusion) provided the highest accuracy in our experiments with the associated database.

In [29-31], quality-dependent experimental fusion results involving fingerprint, face, iris, and signature biometrics are provided.

The rest of our paper is organized as follows: in Section 2, we summarize the characteristics of the fingerprint database we collected & used in our experiments. Section 3 highlights the commercial minutiae extractor / matcher that we utilize. Section 4 presents the proposed matching score fusion systems. Section 5 presents the experimental results, covering these scenarios: (i) traditional (distal phalanx-based) matching & individual phalanx (middle and proximal) matching, (ii) fusing phalanges from the same finger, and (iii) fusing phalanges from multiple fingers. Finally, Section 6 concludes the paper & provides pointers for future work.

## 2. Database Specifications

As there are no publicly available middle and proximal phalanx-based fingerprint image databases we could use, we have collected fingerprint images: (i) from all phalanges (ie. 2 for thumbs, 3 each for other 4 fingers in a hand), (ii) of all 10 fingers, (iii) of 50 individuals (17 females, 33 males, with ages ranging from 23 to 64 years), with 4 impressions per phalanx, in an office environment. The subjects were office workers (e.g., engineers, programmers, students, and managers), hence any professional agricultural work etc. that could affect the image quality adversely is not represented in the database. But we still believe that, this database can form a logical basis for our studies, as we are analyzing to what extent non-distal phalanges can lead to authentication accuracies approaching those of distal ones., ie. our aim is not comparing accuracies originating from a "bad" quality setting (e.g., agricultural work) vs. "good" quality setting (office work), but rather, it is analyzing how incorporating phalanges improves accuracy *within* a setting.

One can also argue that, when distal phalanges are not of sufficient quality to be used in matching, non-distal ones are also "bad", and they may not be used either. But, we may observe that, distal phalanges are generally subject to more wear and tear than non-distal ones, unless the whole hand structure is subject to the same abusive environment (e.g., cement).

The summarized data collection methodology resulted in 500 unique fingers (50 individuals x 10 fingers per individual). We have treated these as 500 different "virtual subjects", as usually done in biometric tests in the literature, in our earlier work [10]. Note that in our current study, for all the analyses that follow, we have excluded database images from thumbs, since they have just a single non-distal phalanx (and all other fingers have two), and we did not want to bias our experimental results with this physical phenomenon. Rather, we have used images from the remaining 4 fingers (ie. index, middle, ring, and little) in a hand, resulting in 400 (50 individuals x 8 fingers per individual) "virtual subjects". For a specific phalanx, impressions 1-2 were obtained in the same day (approximately 10 minutes apart). Impressions 3-4 were obtained in the same day (approx. 10 minutes apart) and these were obtained at least 2 days (on average, 9.5 days, and max. 47 days) *after* impressions 1-2 were obtained.

Fig. 1 shows the labeled phalanges & thenar / hypothenar regions of an individual's right hand, and Fig. 2 shows the image capture setting. An optical, 500 DPI, 480x320 pixel (height x width) fingerprint sensor (FIPS 201/PIV [32] certified) was used during image capture (note that in Fig. 1 & Fig. 2, sensor manufacturer data are masked with black labels to ensure anonymity), that costs approximately 90 US dollars (single unit price, in May 2015).

Sample fingerprint images from a specific finger –acquired in the same day– are given in Fig. 3. We can see that, phalanx 1 image is visually of better quality than other phalanges' images, in terms of ridge clarity. Also, phalanx 2 and 3 images cover larger areas than phalanx 1 image, but they have many horizontal and vertical creases. In the following sections, we elaborate on the effects of these characteristics on minutia extraction & matching performances.

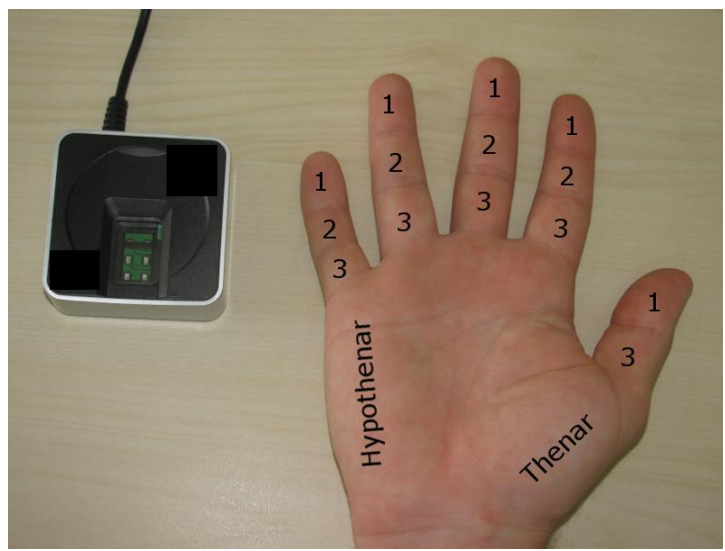

**Fig. 1.** An individual's right hand image with labeled phalanges (1 -distal-, 2 -middle-, and 3 -proximal-), and utilized optical fingerprint sensor

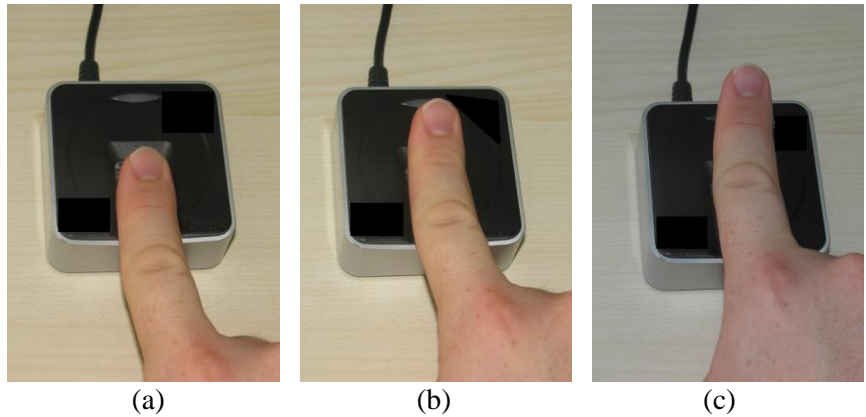

**Fig. 2.** Image capture: (a) phalanx 1 (distal, ie. traditional fingerprint acquisition scenario), (b) phalanx 2 (middle), and (c) phalanx 3 (proximal)

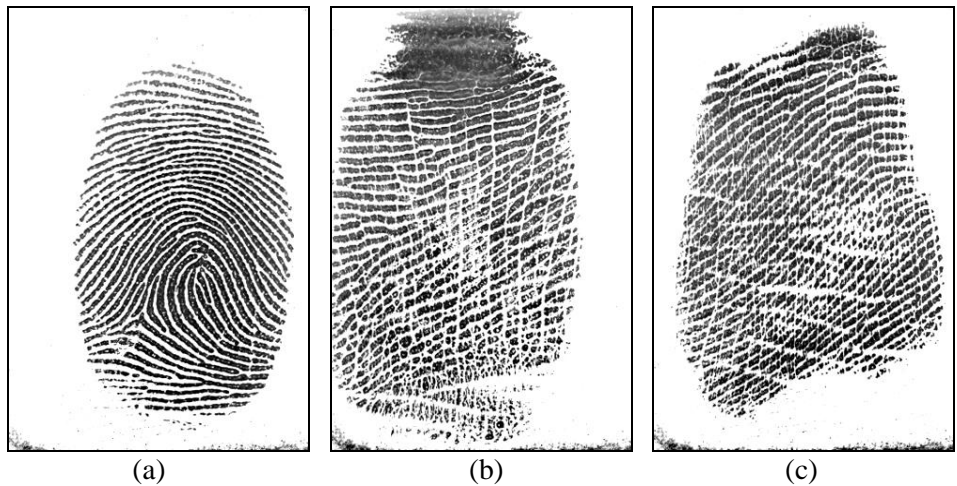

**Fig. 3.** Sample images: (a) phalanx 1, (b) phalanx 2, and (c) phalanx 3 of a unique finger

## 3. Minutia Extractor & Matcher

For extracting the minutiae data ([x,y] coordinates & angle) in ISO/IEC 19794-2:2005 format [1], and for calculating the matching scores originating from phalanx images' minutiae data, we have utilized a commercial minutiae feature extractor / matcher, without any modification, in our experiments. As a result, also considering the fact that a commercial sensor was used without any modification for data collection (cf. Section 2), we are able to analyze the performance of a typical & commercial fingerprint acquisition / feature extraction / matching setup. Note that, for ensuring anonymity, we are not disclosing the vendor ID. But we can note that, this minutiae feature extraction / matching system was among the participants of many public tests & competitions.

In Fig. 4, we show the extracted minutiae overlaid on the associated images, depicted in Fig. 3 (note that just the [x,y] locations of minutiae, without their angle, are shown as red circles for clarity; but the extractor extracts and the matcher uses the minutiae angle information as well). We can see that, there are noisy minutia features present on second and third phalanges.

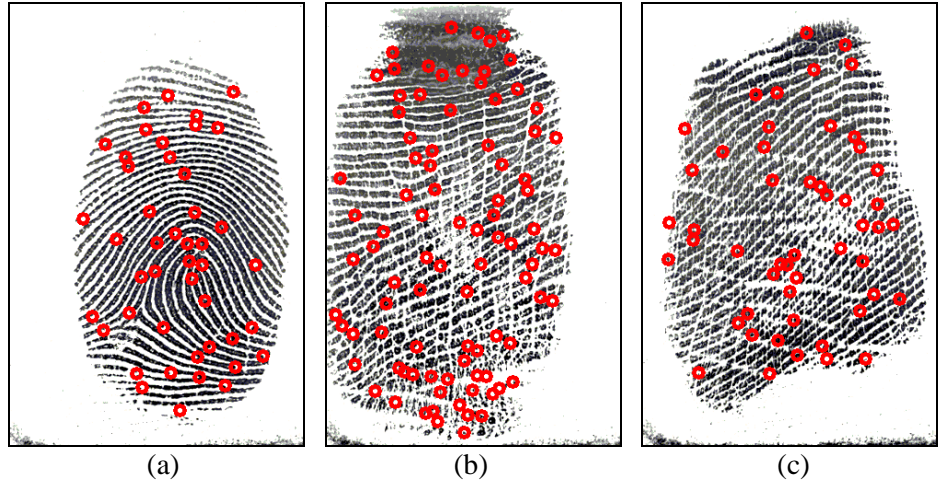

**Fig. 4.** Overlaid minutiae: (a) phalanx 1 (45 minutiae), (b) phalanx 2 (90 minutiae), and (c) phalanx 3 (51 minutiae)

To see if utilized commercial feature extraction & matching system and our optical sensor setup leads to results that are generalizable, we have compared the performance of the extractor & matcher on our (distal) fingerprint database (image size: 480x320 pixels, 500 DPI, 400 unique fingers) with other public (distal) fingerprint databases. For the latter, we have performed minutiae extraction and matching using the same commercial system on FVC 2002 competition's two optical fingerprint databases (namely, DB1A database with image size: 374x388 pixels, 500 DPI, 100 unique fingers, 8 impressions per finger & DB2A database with image size: 560x296 pixels, 569 DPI, 100 unique fingers, 8 impressions per finger) [3]. For this analysis, without any identical template matches, all genuine and imposter matches were performed for all three database scenarios. The associated ROC curves are given in Fig. 5. We can see that the utilized extractor / matcher system performs similarly for these three (distal) fingerprint databases, so we believe that the results that we present in this paper can be generalized (albeit with possibly different authentication performance magnitudes) to other sensor / extractor / matcher combinations as well.

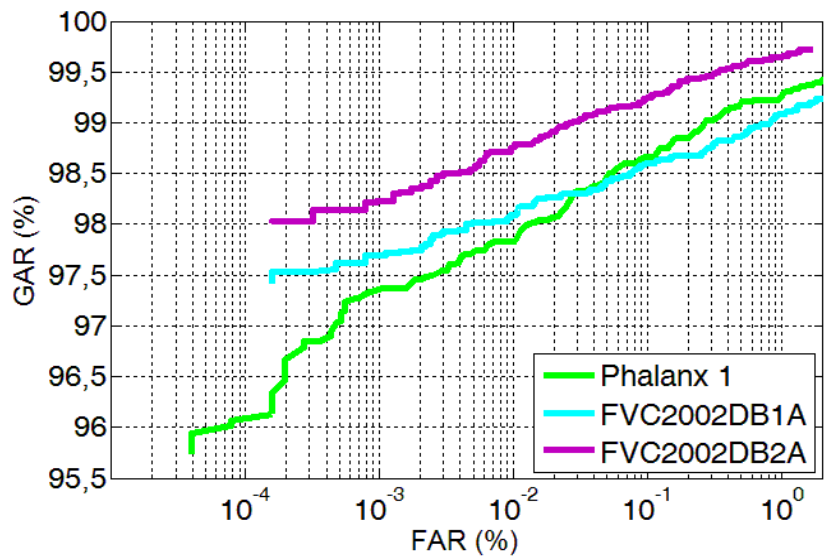

**Fig. 5.** ROC's for phalanx 1 images, and FVC 2002 optical databases' images

We should also note that, the number of unique fingers (virtual subjects) present in our database (ie. 400) is 4 times larger than those of FVC competitions, and depicted ROC curve for phalanx 1 continues towards the lower FAR values (ie. higher security region). As a side note, the reader may notice that in the protocols of FVC competitions [3-5] all imposter scores are not computed, rather, a subset of these scores are computed and used to arrive at the performance figures, such as EER (Equal Error Rate) and FAR1000 (authentication accuracy for 0.1% FAR). In our analyses, we calculated & utilized all imposter scores, for also analyzing the ROC curves around higher security regions of the FAR axis (ie. 0.001% FAR value). The characteristics of ROC curves in such higher security (ie. lower FAR) regions generally contain useful information for applications involving critical infrastructures & large populations (e.g. airport security, border control, AFIS), as opposed to applications with comparatively relaxed security constraints (e.g. gym access).

### 4. Matching Score Fusion Algorithms

For merging scores originating from distal and non-distal phalanges, we have considered three fusion algorithms: (i) NFIQ image-quality-based fusion, (ii) simple sum fusion (that is found by many researchers in biometrics community to be "the best"), and (iii) phalanx-type-based fusion, utilizing static weights for different phalanges. As will be shown in Section 5, the last fusion paradigm led to the best authentication performance results for our database. In this section, we elaborate on these three fusion methodologies.

NFIQ system from National Institute of Standards and Technology (NIST) of U.S. Department of Commerce [12] is a popular fingerprint image quality measurement tool, and we utilized associated quality values for modulating contributions of individual phalanges' images / minutiae templates to the final matching score as explained below. But, as a precursor, we provide measured NFIQ quality (with integer values between 1 -best quality-, and 5 -worst quality-) distributions for individual phalanges, with distinct fingers (Fig. 6 a-c), and with all fingers combined (Fig. 7). We can see that (i) from phalanx 1 to phalanx 3, the overall quality decreases (note that lower values indicate higher quality in NFIQ metric), and (ii) for all phalanges, moving from index finger towards the little finger, the quality generally decreases (ie. higher concentration of quality values towards the value of 5). We should note that, since we collected data without any modification to an existing commercial fingerprint sensor, a more ergonomic (viz. for non-distal phalanges) sensor is likely to increase non-distal image qualities, eliminating ergonomic difficulties some of our subjects had when placing their non-distal phalanges on sensor platen.

In light of these observations, for fusing scores originating from different phalanges, we define:

$$w_{a,b} = q_a + q_b \qquad (1)$$

as the modulating quality value to be used for modulating impression *a*'s and impression *b*'s minutiae matching score, where NFIQ scores (*NFIQ*) are inverted to arrive at integer quality values between 1-5, where 1 indicates lowest quality, and 5 indicates highest quality, as:

$$q_a = 6 - NFIQ_a \qquad (2)$$
$$q_b = 6 - NFIQ_b$$

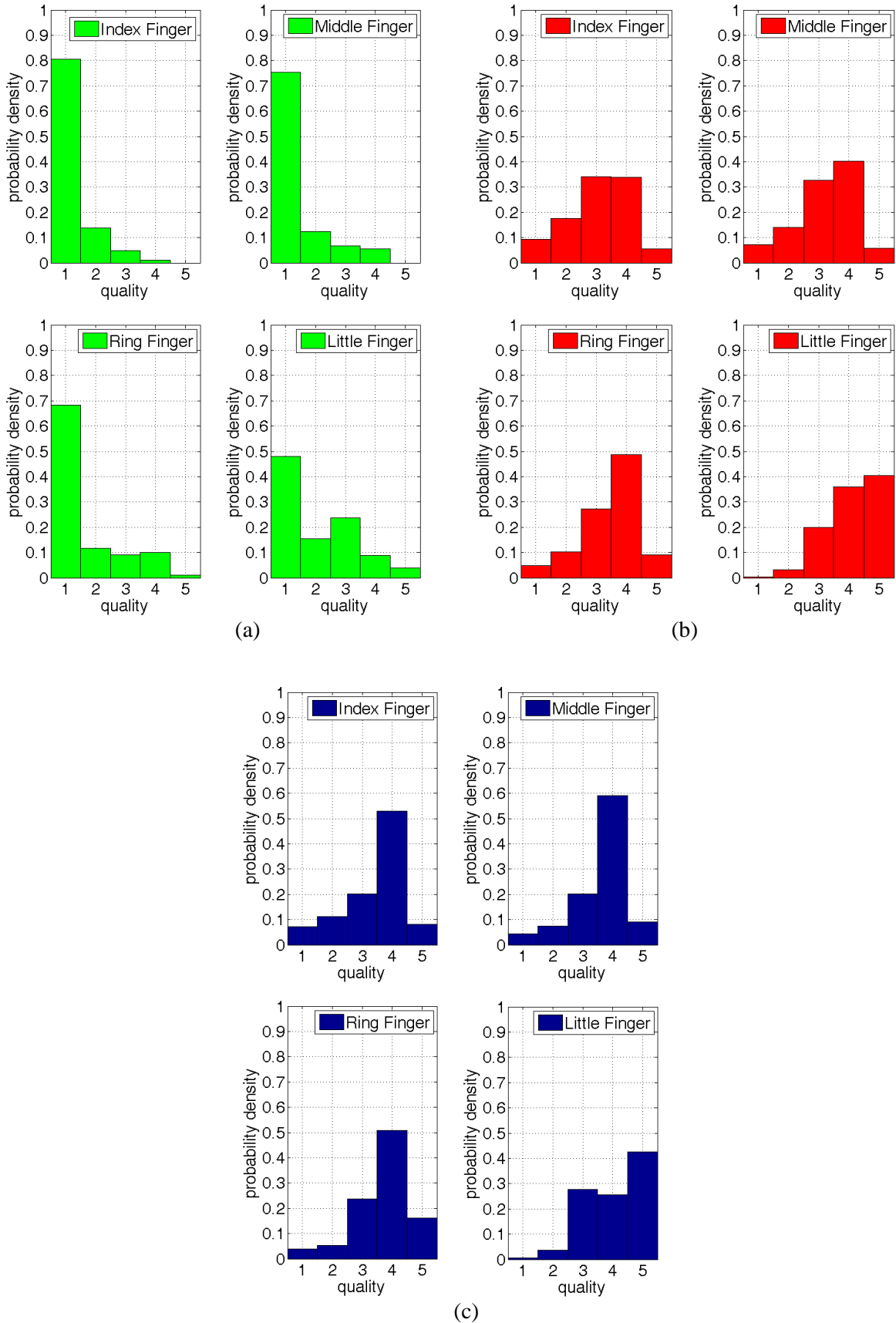

**Fig. 6.** NFIQ quality distributions (distinct fingers): (a) phalanx 1, (b) phalanx 2, and (c) phalanx 3  (1: best quality, 5: worst quality)

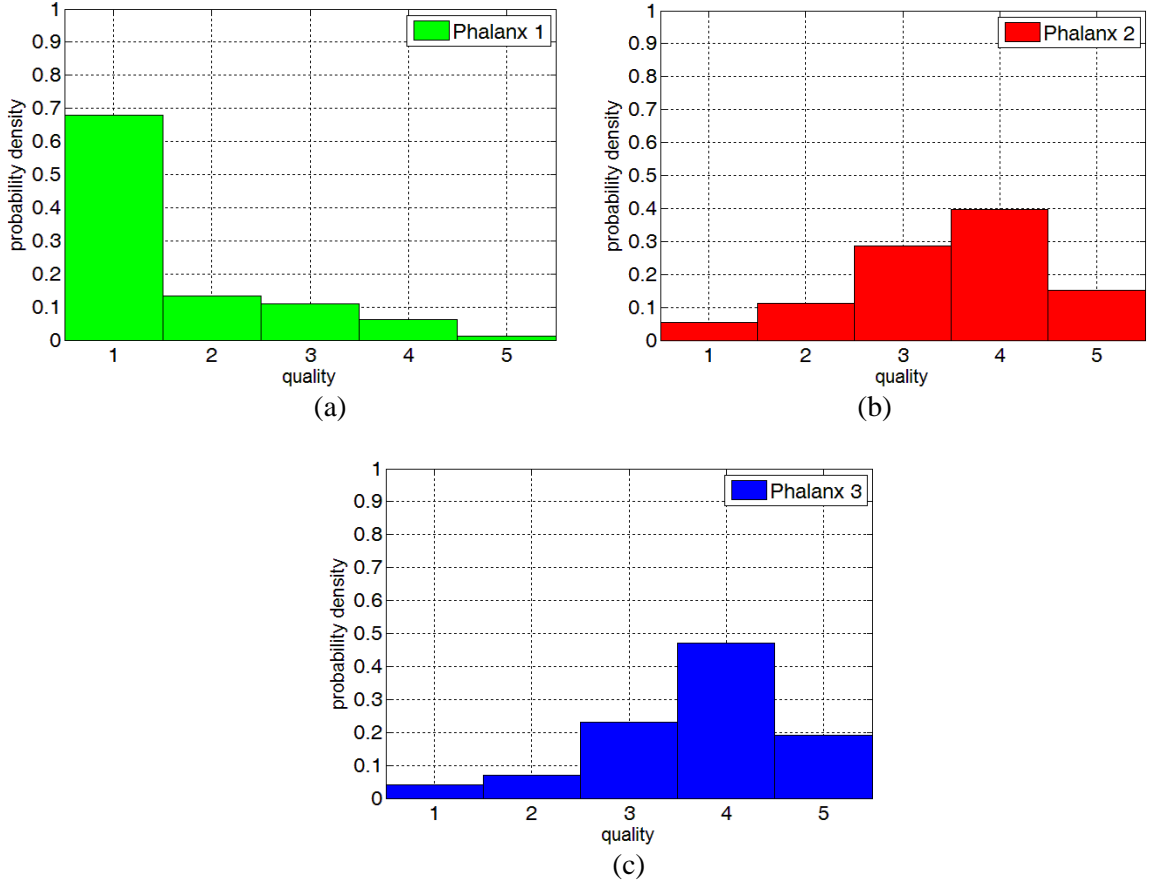

**Fig. 7.** NFIQ quality distributions (all fingers): (a) phalanx 1, (b) phalanx 2, and (c) phalanx 3 (1: best quality, 5: worst quality)

Then, the score for a specific fusion set *M*, $F_M$ is calculated based on matching score $S_{a,b}$ for all *(a,b)* impression pairs in set *M* as:

$$F_M = \frac{\sum_{(a,b) \in M} w_{a,b} S_{a,b}}{\sum_{(a,b) \in M} w_{a,b}} \quad (3)$$

For example, for the scenario of fusing scores from phalanx 2 of index finger with those from phalanx 3 of index finger (an example of *multiphalanx* fusion), the set *M* would include two pairs as *(a,b)*:

(i)  Pair 1:
   - impression *a* from phalanx 2, first impression
   - impression *b* from phalanx 2, second impression

(ii) Pair 2:
   - impression a from phalanx 3, first impression
   - impression b from phalanx 3, second impression

Similarly, for the scenario of fusing scores from phalanx 2 of index and middle fingers, with those from phalanx 3 of index and middle fingers (an example of *multifinger-multiphalanx* fusion) the set *M* would include four pairs as *(a,b)*:

(i) Pair 1:
- impression a from index finger, phalanx 2, first impression
- impression b from index finger, phalanx 2, second impression

(ii) Pair 2:
- impression a from index finger, phalanx 3, first impression
- impression b from index finger, phalanx 3, second impression

(iii) Pair 3:
- impression a from middle finger, phalanx 2, first impression
- impression b from middle finger, phalanx 2, second impression

(iv) Pair 4:
- impression a from middle finger, phalanx 3, first impression
- impression b from middle finger, phalanx 3, second impression

Note that, due to Eqs. 1-2, modulating quality $w_{a,b}$ can have a maximum value of 10 (indicating both impressions had the highest inverted quality value –ie. 5–, meaning increased trust for the associated matching score), and a minimum value of 2 (indicating both impressions had the lowest inverted quality –ie. 1–, meaning decreased trust for the associated matching score), and scores are weighted linearly proportionally with this modulating image quality value.

As the second fusion algorithm, we use simple-sum fusion, where, reusing the parameters (Eqs. 1 & 3) already defined above for image-quality-based fusion, we essentially deal with

$$w_{a,b} = 1 \qquad (4)$$

Further, as the last fusion algorithm, where we assign global static weights as a function of the phalanx type (ie. distal, middle, proximal), we deal with

$$w_{a,b} = \begin{cases} c_1, & \text{if a \& b are from distal phalanx} \\ c_2, & \text{if a \& b are from middle phalanx} \\ c_3, & \text{if a \& b are from proximal phalanx} \end{cases} \qquad (5)$$

where, $0 < c_i < 1$, $\sum c_i = 1$, and $c_3 < c_2 < c_1$. The reason for the last ordering is the relative accuracies of individual phalanges (cf. Fig. 9 in experimental results), where, as expected, distal phalanx provides higher accuracy than middle phalanx, which in turn provides higher accuracy than proximal phalanx. We have experimented with several different sets for these static weights (e.g., $c_1 = 0.7$, $c_2 = 0.2$, $c_3 = 0.1$) and we provide experimental results pertaining to these in Section 5.

Note that, by considering these three fusion algorithms, we have essentially covered:

- Case 1 (*image-quality-based fusion*): All phalanges are weighted adaptively, based on their individual image qualities,
- Case 2 (*simple-sum fusion*): All phalanges are weighted equally,
- Case 3 (*static-weight-based fusion*): Phalanges are weighted with global static weights, which can be thought of as a-priori knowledge, about their respective "reliabilities".

Section 5 includes detailed comparison results regarding these fusion cases.

## 5. Experimental Results

As shown in Section 4, second (middle) and third (proximal) phalanges' images are observed to result in lower qualities with respect those from the first (distal) phalanges. The reasons for this phenomenon could be (i) thicker skin structure on non-distal phalanges, limiting sensor's ability to capture detailed images, (ii) presence of strong horizontal & vertical creases, and (iii) slightly non-traditional usage of the cited commercial sensor with those phalanges, resulting in ergonomic difficulties for some subjects (cf. Fig. 2).

As the fingerprint image quality decreases from phalanx 1 to phalanges 2 & 3, it makes sense to analyze the number of extracted minutiae for these three scenarios. The probability distributions (Fig. 8) show that there is a correlation between the image quality and minutiae numbers: there are more (mainly spurious) minutiae extracted from non-distal phalanges (on average, there were 62.3, 83.4 and 93.9 minutiae for phalanx 1, phalanx 2 and phalanx 3 images, respectively, with standard deviations of 16.0, 23.4 and 30.2, respectively).

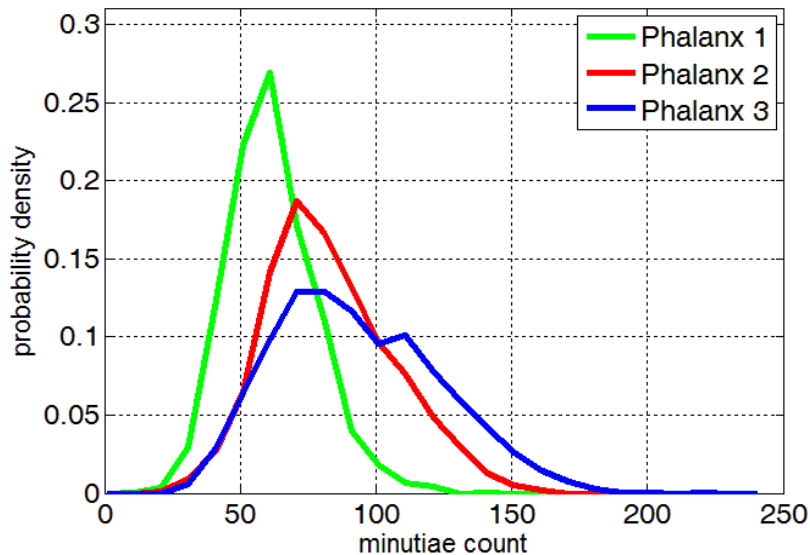

**Fig. 8.** Minutiae number distributions for images from different phalanges

As our first scenario, we compared matching accuracies for distal and non-distal phalanges, without any fusion at all. We will compare the accuracies of the proposed fusion systems, against these baselines, later in this section.

Matching characteristics utilized in Section 3 are exactly followed here (namely, no identical template matches were performed, but, otherwise, all genuine and imposter matches were done). Further, phalanx 2 (middle phalanx) images are only matched with phalanx 2 images, and similarly, phalanx 3 (proximal phalanx) images are only matched with phalanx 3 images and so on. This resulted in 4,800 genuine matching scores, and 2,553,600 imposter matching scores (originating from 400 unique fingers, 1600 distinct images), for each one of these three phalanges. The associated ROC curves are given in Fig. 9. Note that, utilizing all possible genuine and imposter matcher allows the ROC curve to extend beyond $10^{-4}$ % FAR (ie. 1 in 1 million FAR), hence providing accuracy insights about high-security applications (e.g., airport access, critical infrastructure protection).

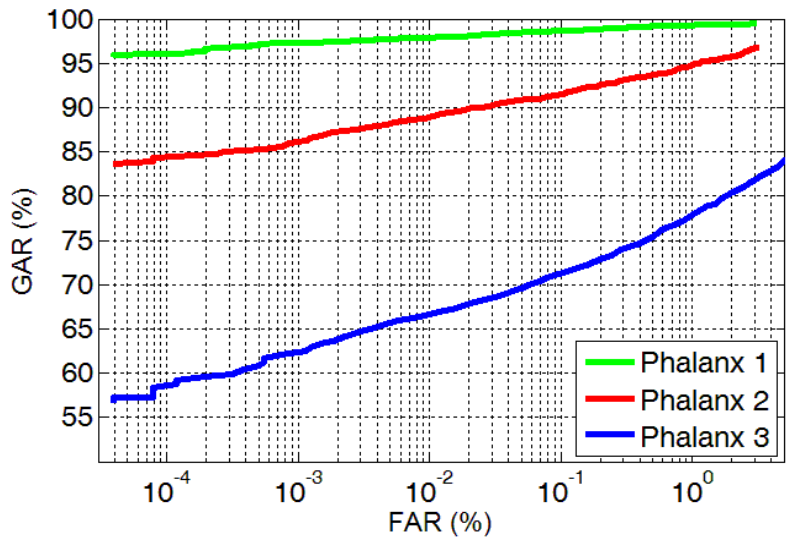

**Fig. 9.** Scenario 1 (no fusion): ROC's for distal and non-distal phalanges

Furthermore, for evaluating statistical significance, we have calculated the 95% confidence intervals for GAR values, corresponding to 0.1%, 0.01%, 0.001% and 0.0001% FAR values (Table 1). Note that, for this and all of the following confidence interval tables, we have used Matlab's *binofit* function for calculating the interval values, based on Clopper-Pearson method.

We can see that, even with the confidence intervals taken into account, there is a performance drop from phalanx 1 to phalanges 2 and 3, as expected. But, phalanx 2 performance (e.g., with 88.9% GAR, at 0.01% FAR) may be considered acceptable, under requisite conditions for some applications, without any fusion at all. We will see the results of fusion below, where phalanx 2 and phalanx 3 fusion leads to performances approaching & exceeding that of phalanx 1.

Within this scenario, in order to see the relative dependencies of matching scores originating from these phalanges, we have calculated the correlation coefficients, R(phalanx_i, phalanx_j), using genuine match scores utilized in Fig. 9, as

$$R(1,2) = R(\text{distal ph., middle ph.}) = 0.39 \quad (6)$$
$$R(1,3) = R(\text{distal ph., proximal ph.}) = 0.31$$
$$R(2,3) = R(\text{middle ph., proximal ph.}) = 0.56$$

which shows that there is a relatively strong positive correlation between scores originating from middle and proximal phalanges.

**Table 1.** Scenario 1 (no fusion): GAR (%): value (V) and 95% confidence intervals (C.I.)

| FAR (%) | Phalanx 1 | | Phalanx 2 | | Phalanx 3 | |
|---|---|---|---|---|---|---|
| | V | C.I. | V | C.I. | V | C.I. |
| 0.1 | 98.7 | [98.3-99.0] | 91.4 | [90.6-92.2] | 71.3 | [70.0-72.6] |
| 0.01 | 97.8 | [97.4-98.2] | 88.9 | [87.9-89.7] | 66.8 | [65.4-68.1] |
| 0.001 | 97.4 | [96.9-97.8] | 86.2 | [85.2-87.2] | 62.3 | [60.9-63.6] |
| 0.0001 | 96.1 | [95.5-96.6] | 84.4 | [83.3-85.4] | 59.0 | [57.6-60.4] |

As our second scenario, we consider *multiphalanx* fusion: fusing (i) phalanges 2 and 3 of a finger, and (ii) all 3 phalanges of a finger, and comparing these with the traditionally considered distal phalanx results. The associated ROC curves given in Fig. 10, and GAR-FAR values with confidence intervals given in Table 2, show that:

(i) combining all three phalanges leads to performances exceeding that of distal phalanx (quite logical to expect), and all three fusion rules lead to similar performances (note that, for static-weight-based fusion, phalanx 1 & phalanx 2 & phalanx 3 weights are experimentally determined as 0.5, & 0.4, & 0.1, respectively),

(ii) fusing phalanges 2 and 3 slightly increases performance with respect to phalanx 2 only scenario, but still could not reach phalanx 1 levels. In this scenario, simple-sum and image-quality based fusion rules lead to quite similar performances, but static-weight-based fusion (with experimentally determined phalanx 2 & phalanx 3 weights of 0.7 & 0.3, respectively) exceed their performance.

The reason for the similar performances of simple-sum and image-quality based fusion rules may be the tendency of the utilized fingerprint image quality measurement tool (ie. NFIQ) in measuring the qualities of distal phalanges (for which it is originally designed & trained) with higher reliability than those of non-distal phalanges.

For determining the cited static weights, we have experimented with several weight sets, and found the best-performing one: for fusing phalanges 1, 2 and 3, the associated ROC curves given in Fig. 11 are obtained. For fusing phalanges 2 and 3, the associated ROC curves given in Fig. 12 are obtained. Based on these curves, the weight values cited before are determined.

As our third and final scenario, we consider *multifinger-multiphalanx* fusion: (a) firstly, fusing 2 fingers and 2 non-distal phalanges (ie. total of 4 phalanges) and comparing these with the distal phalanx results, (b) secondly, fusing 2 (ie. total of 4 non-distal phalanges), 3 (ie. total of 6 non-distal phalanges) and 4 (ie. total of 8 non-distal phalanges) fingers, and comparing these with the distal phalanx results. Note that from this point on, we have only evaluated static-weight-based fusion (which led to performances superior to image-quality-based and simple-sum fusion rules as shown above), with the optimal weight sets found before.

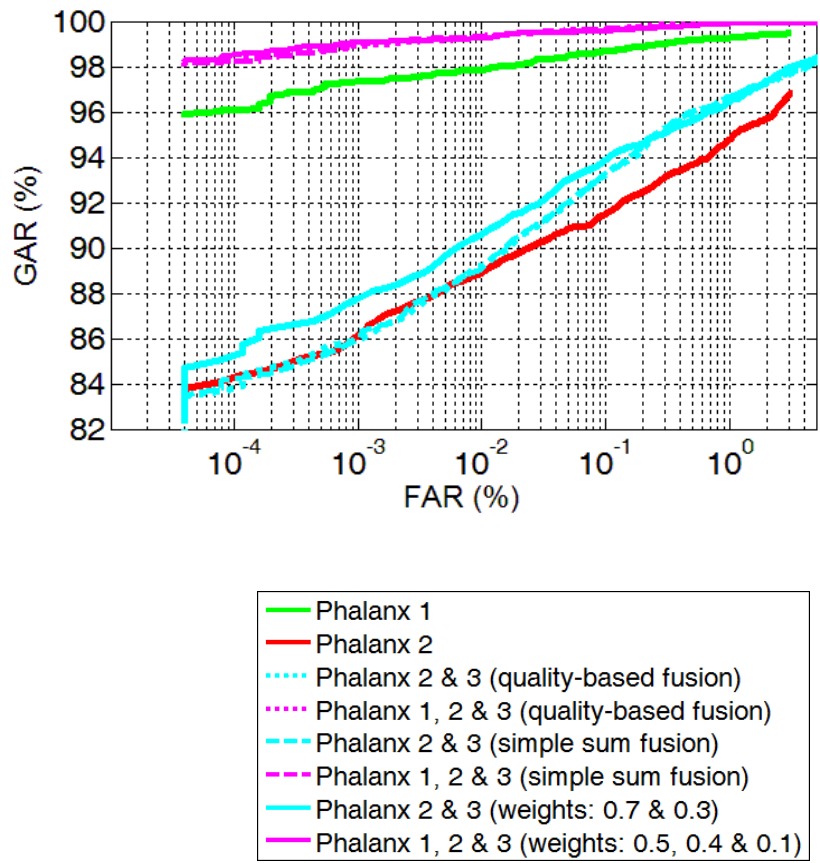

**Fig. 10.** Scenario 2: ROC's for distal (Phalanx 1), middle (Phalanx 2) phalanges and multiphalanx fusion

Table 2. Scenario 2 (multiphalanx fusion): GAR (%): value (V) and 95% confidence intervals (C.I.)

| FAR (%) | Phalanx 1+2+3 fusion (Image-quality) | | Phalanx 1+2+3 fusion (Simple sum) | | Phalanx 1+2+3 fusion (Static-weight) | |
|---|---|---|---|---|---|---|
| | V | C.I. | V | C.I. | V | C.I. |
| 0.1 | 99.7 | [99.4-99.8] | 99.6 | [99.4-99.8] | 99.6 | [99.3-99.7] |
| 0.01 | 99.3 | [99.0-99.5] | 99.3 | [99.1-99.6] | 99.3 | [99.0-99.5] |
| 0.001 | 98.9 | [98.6-99.2] | 99.0 | [98.7-99.3] | 99.1 | [98.7-99.3] |
| 0.0001 | 98.3 | [97.9-98.6] | 98.2 | [97.8-98.6] | 98.5 | [98.1-98.8] |

| FAR (%) | Phalanx 2+3 fusion (Image-quality) | | Phalanx 2+3 fusion (Simple sum) | | Phalanx 2+3 fusion (Static-weight) | |
|---|---|---|---|---|---|---|
| | V | C.I. | V | C.I. | V | C.I. |
| 0.1 | 93.3 | [92.5-94.0] | 93.3 | [92.6-94.0] | 93.9 | [93.2-94.6] |
| 0.01 | 89.2 | [88.3-90.1] | 89.1 | [88.2-90.0] | 90.6 | [89.7-91.4] |
| 0.001 | 86.0 | [85.0-87.0] | 86.0 | [85.0-87.0] | 87.8 | [86.8-88.7] |
| 0.0001 | 83.8 | [82.8-84.9] | 84.2 | [83.2-85.2] | 85.2 | [84.2-86.2] |

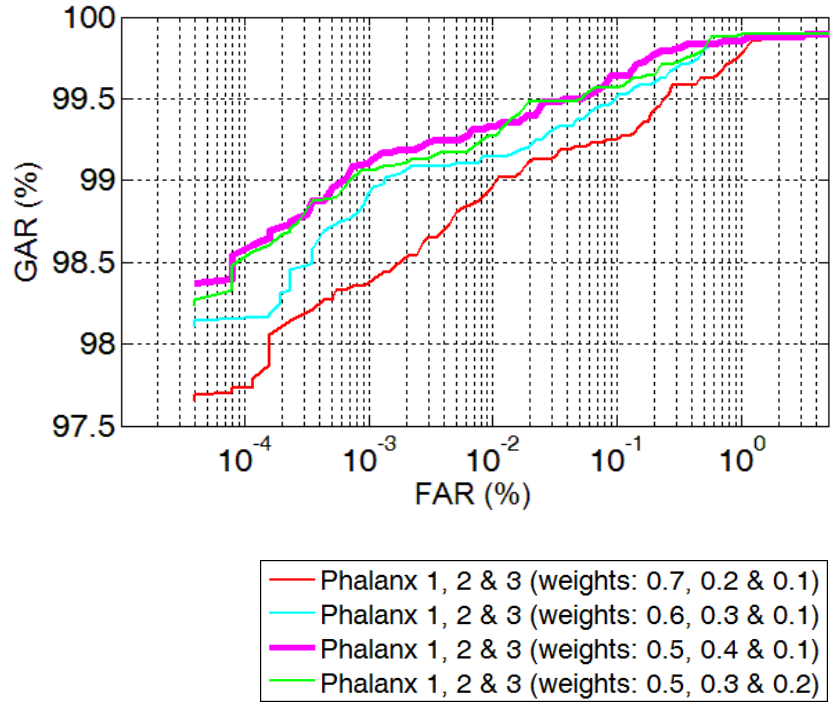

**Fig. 11.** Scenario 2: ROC's for multiphalanx (Phalanges 1&2&3) fusion: evaluated static weight sets

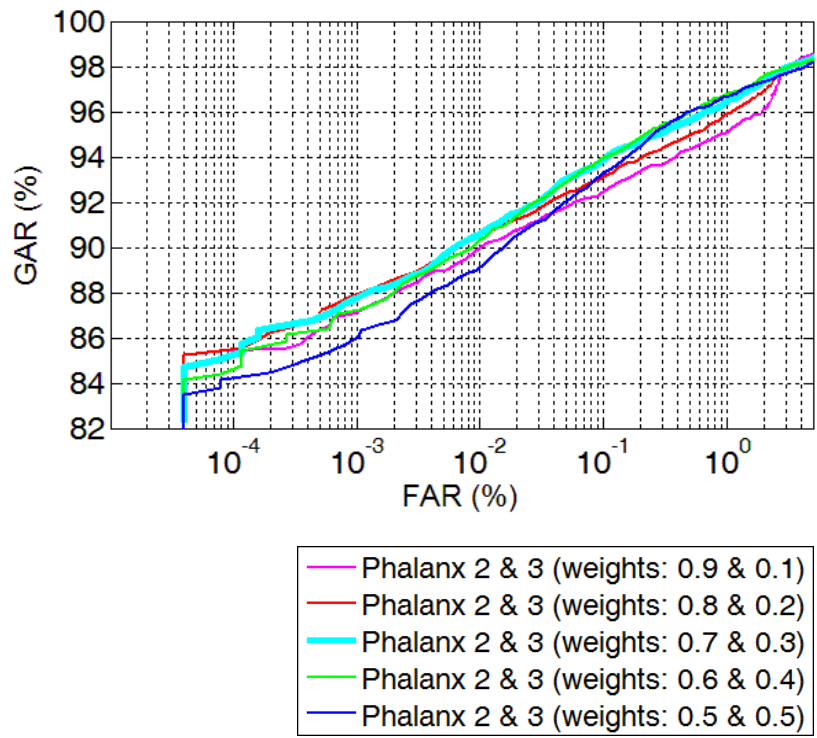

**Fig. 12.** Scenario 2: ROC's for multiphalanx (Phalanges 2&3) fusion: evaluated static weight sets

For the first case, associated ROC curves in Fig. 13 and GAR-FAR values in Table 3 show that fusing 2. and 3. phalanges of two fingers with the highest relative qualities, namely index and middle fingers (cf. Fig. 6), leads to performances exceeding those of distal phalanges. As a result, this may point at a feasible fusion system, that can be utilized if phalanx 1 images are not usable, e.g. due to low quality –distal– finger surfaces or missing digits.

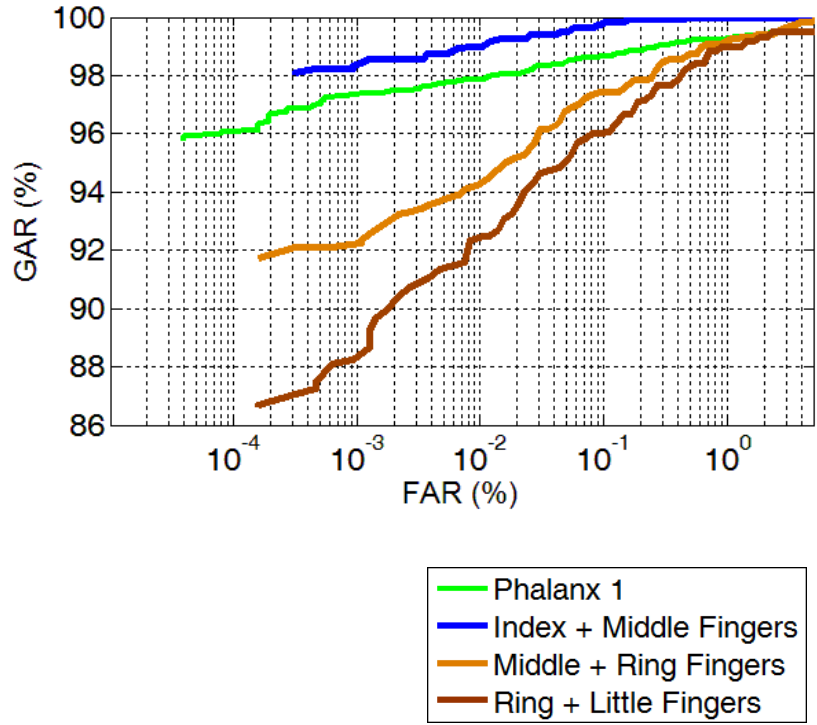

**Fig. 13.** Scenario 3a: ROC's for two-phalanx, two-finger fusion (static-weight-based) compared to distal phalanx

**Table 3.** Scenario 3a: GAR (%): value (V) and 95% confidence intervals (C.I.)

| FAR (%) | Phalanx 2+3 fusion index+middle fingers | | Phalanx 2+3 fusion middle+ring fingers | | Phalanx 2+3 fusion ring+little fingers | |
|---|---|---|---|---|---|---|
| | V | C.I. | V | C.I. | V | C.I. |
| **0.1** | 99.8 | [99.4-100] | 97.4 | [96.4-98.2] | 96.0 | [94.8-97.1] |
| **0.01** | 99.0 | [98.3-99.5] | 94.3 | [92.8-95.5] | 92.5 | [90.8-93.9] |
| **0.001** | 98.4 | [97.5-99.0] | 92.2 | [90.6-93.7] | 88.3 | [86.4-90.1] |

Note that, in this scenario, 4 phalanx images from an individual are needed for authentication, hence we can say that data acquisition requires 4 times longer than that of acquiring a traditional (distal) phalanx image. This is obviously larger than traditional scenario, but it is expected to be on the order of few seconds, and it may be considered feasible for specific applications, under requisite conditions. As evident in Fig. 13, combining middle + ring, and ring + little fingers, which entails the same acquisition overhead (ie. 4 phalanges), does not provide performances comparable with that of distal phalanges.

For the second case, we wanted to analyze if adding more fingers increase accuracy even further (obviously, at the cost of increased data acquisition time). We can see in Fig. 14 and Table 4 that, slight accuracy increases especially for the higher security (ie. lower FAR) region of the ROC's are indeed present with these additions. But these minute accuracy increases (which are only possible with the further acquisition time increases) may be considered unnecessary for many applications.

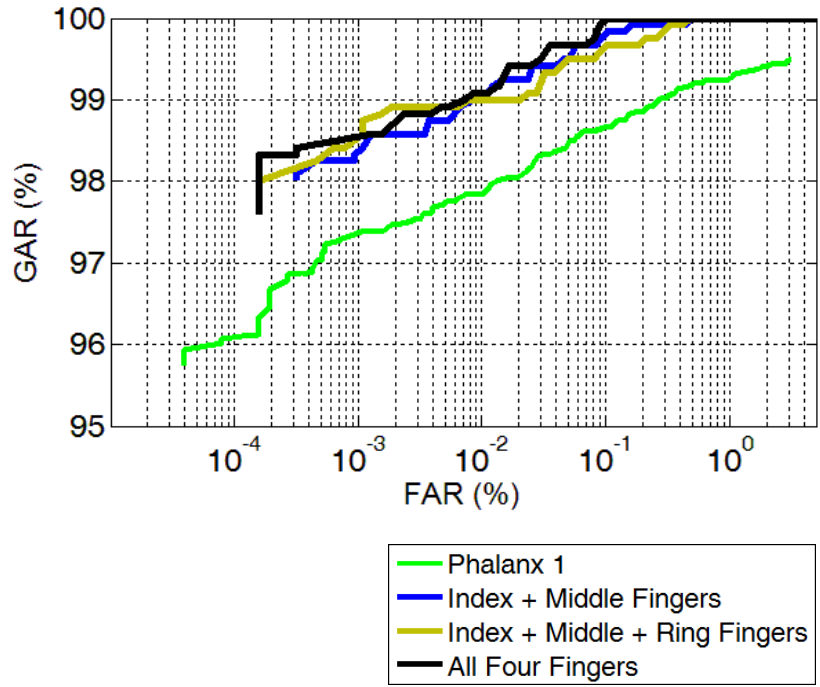

**Fig. 14.** Scenario 3b: ROC's for two-phalanx & (two-finger, three-finger, four-finger) fusion (static-weight-based) compared to distal phalanx

**Table 4.** Scenario 3b: GAR (%): value (V) and 95% confidence intervals (C.I.)

| FAR (%) | Phalanx 2+3 fusion *index+middle fingers* | | Phalanx 2+3 fusion *index+middle+ring fingers* | | Phalanx 2+3 fusion *index+middle+ring+little fingers* | |
|---|---|---|---|---|---|---|
| | V | C.I. | V | C.I. | V | C.I. |
| 0.1 | 99.8 | [99.4-100] | 99.6 | [99.1-99.9] | 100 | [99.7-100] |
| 0.01 | 99.0 | [98.3-99.5] | 99.0 | [98.3-99.5] | 99.1 | [98.4-99.5] |
| 0.001 | 98.4 | [97.5-99.0] | 98.5 | [97.7-99.1] | 98.5 | [97.7-99.1] |

## 6. Conclusions and Future Work

Motivated by the fact that fingerprints continue to be the dominant biometric modality across different applications and geographies (mainly due to cost advantages, sensor sizes, standardization & interoperability benefits), and traditionally used distal phalanx section of fingers being unusable (due to manual work, wear & tear, missing digits etc.) for a portion of the population, we proposed a matching score fusion system for non-distal phalanges. Evaluating simple-sum, image-quality, and phalanx-type based fusion rules, and utilizing a medium-size database (consisting of 50 individuals, 400 unique fingers, 1600 distinct images) collected in our laboratory with a commercial low-cost optical fingerprint sensor, and a commercial minutiae extractor / matcher system without any modification, we simulated a traditional commercial fingerprint authentication scenario.

Our experimental results show that simple-sum and image-qualiy-based fusion rules lead to similar performances, and the phalanx-type-based fusion rule with global static weights outperforms these. We further showed that fusing index and middle fingers' non-distal phalanges (ie. utilization of 4 distinct phalanges) leads to authentication performances *exceeding* those of distal-only phalanges. Furthermore, when all three phalanges are utilized, we also obtain significant performance improvements with respect to distal-phalanx-only scenario.

Detailed analyses regarding the number of extracted minutiae, image quality distributions as a function of finger & phalanx types, ROC curves and GAR-FAR tables with calculated statistical confidence intervals for several experimental scenarios including traditional and proposed fusion settings, highlight the characteristics of fingerprint phalanx-based matching (first one in the reported literature for commercial sensor / feature extractor / matcher paradigm).

The proposed fusion system may be a viable alternative under requisite conditions, for cases where (i) distal fingerprint images are not usable, and (ii) moving to a new modality (e.g., iris) is not possible due to economical & infrastructure limits, since the existing fingerprint matching paradigm (sensor / feature extractor / matcher) is enriched with the proposed fusion scheme, rather than replacing it as a whole.

Also, when all phalanges are considered usable, proposed fusion system increases performance compared to the traditional distal matching case, as expected.

Increasing the database size is a possible future work (but note that, current database size provides statistical significances exceeding those of many established public tests and associated literature). Another research area could be designing and testing a pre-processor for removing many creases present on non-distal phalanges, and running minutiae feature extractor & matcher after this pre-processing step, to see its effect on matching performances.

When a database that contains statistically large numbers of accurately labeled *latent* fingerprint phalanx images (collected from crime scenes etc.) is made available to the research community, the outlined experiments can be repeated to see the effects of proposed fusion algorithms on latent images. Note that, unfortunately, currently available latent fingerprint databases (e.g. [33]) do not satisfy these constraints.

Another possible future work could be *manually* marking the minutiae features on fingerprint non-distal phalanx images and analyzing the matching performance differences between automatically extracted & manually extracted minutiae data. A study involving widely used (distal) fingerprint databases that quantified such performance differences can be found in [34].